\documentclass[letterpaper, 10 pt, conference]{ieeeconf}

\IEEEoverridecommandlockouts 
\usepackage{amsmath}                      
\usepackage{amssymb}                      
\usepackage{algorithm}                    
\usepackage{algpseudocode}
\usepackage[T1]{fontenc}                  
\usepackage{graphicx}
\usepackage{subcaption}                  
\usepackage[urlcolor=blue,colorlinks=true]{hyperref} 
\usepackage[utf8]{inputenc}               
\usepackage{soul}                         
\usepackage[usenames,dvipsnames]{xcolor}  
\usepackage{multirow}                     
\usepackage{makecell}					  
\usepackage[font=footnotesize]{caption}
\usepackage[letterpaper]{geometry}
\usepackage{fancyhdr}
\usepackage{afterpage} 

\newcommand{\ceil}[1]{\left\lceil #1 \right\rceil}

\fancypagestyle{firstpage}{
  \fancyhf{}
  
  \geometry{
    top=60pt,
    left=48pt,
    right=48pt,
    bottom=43pt
  }
}

\newcommand{\dof}{{\sc dof}}
\newcommand{\dofs}{{\sc dof}s}
\newcommand{\cspace}{\ensuremath{\mathcal{C}_{space}}}

\newcommand{\env}{\mathcal{E}}
\newcommand{\robots}{\mathcal{R}}
\newcommand{\query}{\mathcal{Q}}
\newcommand{\paths}{\mathcal{P}}
\newcommand{\conflict}{\mathcal{C}}

\newcommand{\algorithmicinput}{\textbf{Input:}}
\newcommand{\algorithmicoutput}{\textbf{Output:}}
\newcommand{\INPUT}{\item[\algorithmicinput]}
\newcommand{\OUTPUT}{\item[\algorithmicoutput]}

\algdef{SE}[DOWHILE]{Do}{doWhile}{\algorithmicdo}[1]{\algorithmicwhile\ #1}%

\makeatletter
\renewcommand{\ALG@beginalgorithmic}{\small}
\makeatother

\algrenewcommand\algorithmicindent{1em} 

\begin{document}

	\title{\LARGE \bf
		K-ARC: Adaptive Robot Coordination for Multi-Robot Kinodynamic Planning }
	
	\author{Mike Qin$^{1}$, Irving Solis$^{1,2}$, James Motes$^{1}$, Marco Morales$^{1,3}$, and
		Nancy M. Amato$^{1}$
		\thanks{$^{1}$  University of Illinois Urbana-Champaign, Siebel School of Computing and Data Science, Urbana, IL, 61801, USA. \{yudiqin2, jmotes2, moralesa, namato\}@illinois.edu.}
        \thanks{$^{2}$  Texas A\&M University, Department of Computer Science and Engineering, College Station, TX,
			77840, USA.  \{irvingsolis89\}@tamu.edu.}%
		\thanks{$^{3}$Instituto Tecnológico Autónomo de México (ITAM),
            Department of Computer Science, México City, 01080, México. \{marco.morales\}@itam.mx. Supported in part by Asociación Mexicana de Cultura A.C.}%
	}
	
	\maketitle
	
	\thispagestyle{empty} 
	\pagestyle{plain}     

\begin{abstract}


This work presents Kinodynamic Adaptive Robot Coordination (K-ARC), a novel algorithm for multi-robot kinodynamic planning. Our experimental results show the capability of K-ARC to plan for up to 32 planar mobile robots, while achieving up to an order of magnitude of speed-up compared to previous methods in various scenarios. K-ARC is able to achieve this due to its two main properties. First, K-ARC constructs its solution iteratively by planning in segments, where initial kinodynamic paths are found through optimization-based approaches and the inter-robot conflicts are resolved through sampling-based approaches. The interleaving use of sampling-based and optimization-based approaches allows K-ARC to leverage the strengths of both approaches in different sections of the planning process where one is more suited than the other, while previous methods tend to emphasize on one over the other. Second, K-ARC builds on a previously proposed multi-robot motion planning framework, Adaptive Robot Coordination (ARC), and inherits its strength of focusing on coordination between robots only when needed, saving computation efforts. We show how the combination of these two properties allows K-ARC to achieve overall better performance in our simulated experiments with increasing numbers of robots, increasing degrees of problem difficulties, and increasing complexities of robot dynamics.

\end{abstract}
		
\section{Introduction}

Multi-robot systems have gained significant usage in real-world applications including warehouse automation, delivery, and surveillance tasks. Thus, it is important to develop scalable and efficient multi-robot kinodynamic planning (MRKP) algorithms that can compute trajectories for a team of robots that are collision-free and satisfy the robots' dynamic constraints. 

MRKP is an instance of the more general problem of multi-robot planning, which can typically be approached in three ways. First, coupled methods treat the team of robots as one single entity, where the states of all the robots are combined and the problem can then be solved by any single-robot planner. Coupled methods are able to produce plans where high levels of coordination between the robots are needed but can often fail as the number of robots increases, due to the exponential growth of the search space, which makes finding a feasible solution very computationally expensive. Second, decoupled methods plan for robots individually and separately, where the plan for each robot is generated sequentially. Decoupled methods are typically faster in simpler problems but can fail when high levels of coordination between the robots are needed. This is because decoupled methods cannot adjust paths that are already found, leading to situations where robots may encounter deadlocks that cannot be resolved without completely re-planning their paths. To address MRKP, a naive coupled or decoupled adaptation of Kinodynamic Rapidly Exploring Random Trees (RRT) \cite{lavalle-kino-rrt} can be used. 

Hybrid methods, on the other hand, leverage the strengths of both coupled and decoupled methods. Most existing work can be classified as hybrid and operates in a hybrid manner and builds on Conflict-Based Search (CBS) \cite{SHARON201540}, a well-known algorithm for solving multi-agent pathfinding, as an underlying multi-robot framework. These methods differ in their specific planners used for MRKP, but are the same in the high-level structure, which consists of two planning layers, where a high-level planner manages the conflicts between robots, and a low-level planner directly plans individual robot paths \cite{Moldagalieva} \cite{Juncheng} \cite{10611078} \cite{Kottinger}.  

Our previous work, adaptive robot coordination (ARC) \cite{solis-arc} can also be classified as a hybrid algorithm. ARC instead solves the multi-robot motion planning (MRMP) problem, a simplified version of MRKP that does not consider robot dynamics. The core idea of ARC is to dynamically adjust the dimensions of the problems, including the environment size and the number of robots involved, based on the levels of coordination needed between the robots, making it an efficient algorithm in many different scenarios. In our original work, we only assumed holonomic robots and did not consider dynamic constraints on the robots, which puts restrictions on how the robots are allowed to move. As we will discuss later, our original work is insufficient when this assumption is relaxed.

Here, we present Kinodynamic Adaptive Robot Coordination (K-ARC), an efficient, hybrid multi-robot kinodynamic planner. In K-ARC, we find the initial individual path for each robot without taking into account inter-robot collisions and dynamic constraints. Instead of resolving inter-robot collisions iteratively on the entire paths, we operate on segments of the paths. We first divide the paths into an equal number of segments. In each segment, we find individual kinodynamically-feasible paths through optimization-based approaches. These paths are then checked for conflicts, where conflicts are resolved through sampling-based approaches. K-ARC strategically uses optimization-based and sampling-based approaches where each is better suited, which we show significantly reduces computation time.


In summary, our contributions are:

\begin{itemize}
    \item A novel MRKP algorithm, K-ARC, that adapts a ARC-styled subproblem-based multi-robot conflict resolution, to consider robot dynamics. 
    \item A novel framework that employs a hybrid approach that leverages the strengths of both sampling-based and optimization-based planning at different levels of problem complexity and size.
    \item An experimental analysis that shows K-ARC's ability to plan for up to 32 robots while achieving one to two magnitudes of speed-up across different scenarios compared to baseline methods from recent literature.
\end{itemize}


\section{Preliminaries and Problem Formulation}
In this chapter, we introduce preliminaries and relevant work, and we formally define the problem that we study.

\subsection{Motion Planning}
Motion planning is the problem of finding a valid continuous path for a robot from a start to a goal configuration within the configuration space (\cspace)~\cite{lozano-plan}, the set of all possible robot configurations. A configuration $c \in \cspace$ encodes the robot's degrees-of-freedom (\dof), which fully parameterizes the robots' geometric pose. As the number of {\dofs} in a motion planning problem increases, representing the \cspace \hspace{1mm} explicitly becomes intractable~\cite{canny-plan}. This results in the development of sampling-based methods, where the main idea is to approximate the \cspace \hspace{1mm} using a graph, i.e., a roadmap \cite{kavraki-prm} or a tree \cite{lavalle-rrt}, rather than explicitly constructing the \cspace. 

\subsection{Kinodynamic Planning}

Kinodynamic planning takes the problem of motion planning and accounts for dynamic constraints on the movements of the robots, where the differential constraint $x' = f(x, u)$ dictates how the robots are allowed to move. Here, $x$ is the state variable that encompasses the configuration ($c \subseteq x$) and its time derivatives, and $u$ is the control variable. 

While kinodynamic planning can be considered as a subproblem of motion planning, we use the term motion planning to strictly refer to the problem of kinematic motion planning, where robot dynamics are not taken into account, and to distinguish it from kinodynamic planning in this work. 

Kinodynamic planning can be solved directly through sampling-based methods. Existing work consists of different variants of RRT \cite{lavalle-kino-rrt} \cite{karaman} \cite{webb-kino-rrt*} \cite{hauser-kino-rrt}, either with or without optimality guarantees. 

Kinodynamic planning can also be solved through optimization-based methods \cite{chomp} \cite{stomp} \cite{schulman}. Instead of explicitly planning in the \cspace, optimization-based approaches model the problem as a constrained optimization problem, where the obstacles and robot dynamics are explicitly defined as either part of the cost function or added as constraints to the optimizer. 

\subsection{Multi-Robot Kinodynamic Planning}
In this work, we address the problem of multi-robot kinodynamic planning, which we formally define here. 

Given a team of $ n \in \mathbb{N}$ homogeneous robots $R = \{r_1, r_2, ..., r_n\}$ operating in a shared workspace $\mathcal{W} \subseteq \mathbb{R}^d$ where $d \in \{2, 3\}$, our goal is to find dynamically feasible, collision-free trajectories for each robot from their respective start state to a goal region. Each robot $r_i$ has state $x_i \in X_i $ and control input $u_i \in U_i $ and its motion is governed by the following continuous-time dynamics:
\[
\dot{x}_i = f(x_i, u_i)
\]
The dynamics are discretized over time with a fixed timestep $\Delta t$. During each timestep, the control input is assumed to be constant, and the state transition for each robot at timestep $k$ is given by the following Euler integration:
\[
\ x_{i,k+1} = x_{i,t} + f(x_{i,k}, u_{i,k}) \Delta t
\]
More formally, our problem is formulated as:

Find a trajectory \( \{x_{i,1}, x_{i,2}, \dots, x_{i,T}\} \) and control inputs \( \{u_{i,1}, u_{i,2}, \dots, u_{i,T-1}\} \) for each robot \( r_i \in R \) 
\begin{equation}
    \label{objective}
    \begin{aligned}
        & \text{subject to:} \\
        &      x_{i,1} = x_{i,s} \quad \text{(1)}\\
        &      x_{i,T} \in X_{i,g} = \{x \in X \mid \|x - x_g\| \leq \alpha\} \quad \text{(2)}\\
        &      x_{i, k+1} = x_{i,k} + f(x_{i,t}, u_{i,t}) \Delta t \quad \forall k \in \{1, 2, ..., T - 1\} \quad \text{(3)}\\ 
        &      u_{i,k} \in U \quad \text{(4)} \\
        &      c_{i,k} \in \mathcal{W}_{\text{free}} \text{ where } c_{i, k} \subseteq x_{i, k} \quad \text{(5)}\\
        &      \|c_{i,k} - c_{j,k}\| \geq d_{\text{min}}, \quad \forall i \neq j \quad \text{(6)} 
    \end{aligned}
\end{equation}

In this formulation, (1) (2) are the start and goal constraints, respectively, and (3) corresponds to the dynamic constraints. In (4), we can only choose the control inputs from a set of allowable control inputs. In (5), we require the configuration of a robot's state to be within free space. In (6) we require the configurations of any two robots to be at least a certain distance away at any timestep.

\section{Related Work}
In this section, we give a brief overview of current approaches on multi-robot kinodynamic planning. We also formally introduce our previous work on multi-robot motion planning, ARC, along with experiment evaluations, and discuss why it cannot be directly extended to solve multi-robot kinodynamic planning. 

\subsection{Multi-Robot Kinodynamic Planning}

While the problem of multi-robot kinodynamic planning can be intuitively solved by planning in the joint space of all robots, such an approach does not scale well. Similar to single-robot kinodynamic planning, multi-robot kinodynamic planning can be addressed by either sampling-based methods, optimization-based methods, or a mix of both, and to deal with increasing dimensionality of the state space, existing work tries to decouple the space. 

On the sampling-based side, the work in \cite{Kottinger} extends the Conflict-Based Search (CBS) algorithm \cite{SHARON201540} for multi-agent path finding to MRKP and adds kinodynamic constraints implicitly in the planning stage. Another work in \cite{Cain} extends the RRT* \cite{webb-kino-rrt*} algorithm for optimal motion planning by planning for the robots sequentially where each robot's trajectory is treated as a dynamic obstacle when planning for subsequent robots.

Other methods use a mix of search, sampling, and optimization based approaches. The work in \cite{Juncheng} first finds the discrete paths for the robots and uses a decoupled optimization algorithm for postprocessing the paths into kinodynamically feasible trajectories. Another work in \cite{ChenAAAI21s2m2} uses a Mixed-Integer Linear Programs (MILP) for finding a collision-free path for each robot and applies a priority-based search to coordinate the robots. A more recent work in \cite{Moldagalieva} adapts a discontinuity-bounded single robot planner \cite{honig} for multi-robot and applies trajectory optimization as a postprocessing step for fixing the discontinuities to ensure the trajectories' kinodynamic feasibility.

\subsection{Adaptive Robot Coordination}
In ARC, we assume the robots are holonomic and do not consider robots' dynamic constraints. Intuitively, the same framework can be applied directly in solving a kinodynamic problem. However, since the local subproblem is also a kinodynamic problem, we need to define a tolerance for the local goal. Doing so, however, could result in a goal state that is off from the goal state we defined, which then leaves a discontinuity between the local goal state and the start state of the rest of the path. To resolve this issue, we could first try replanning for the two states. Since a straight line is not feasible, we'd have to then run a solver to connect the two states. The solver might take us closer to the goal, but then again, it cannot guarantee to find the exact goal state. Alternatively, we could try replanning from the state that we found to the global goal. This is feasible but not efficient, since we'd have to do this for every conflict, and the effort of our original solution is essentially wasted. 

To tackle this problem, we propose to construct the solutions iteratively, as opposed to making modifications directly on the solutions (as in ARC). We explain our approach in detail in the following section.

\section{Method}
\begin{figure*}[t] 
\footnotesize
    \centering
    \begin{subfigure}[b]{0.245\linewidth}
        \includegraphics[width=1.0\linewidth]{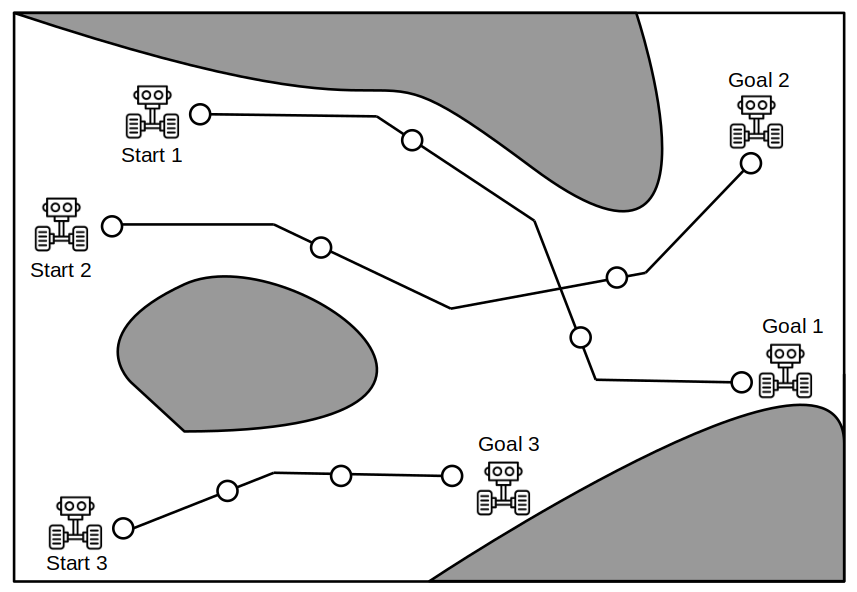}
        \caption{\footnotesize\vspace{0mm}}
    \end{subfigure}
    \begin{subfigure}[b]{0.245\linewidth}
        \includegraphics[width=1.0\linewidth]{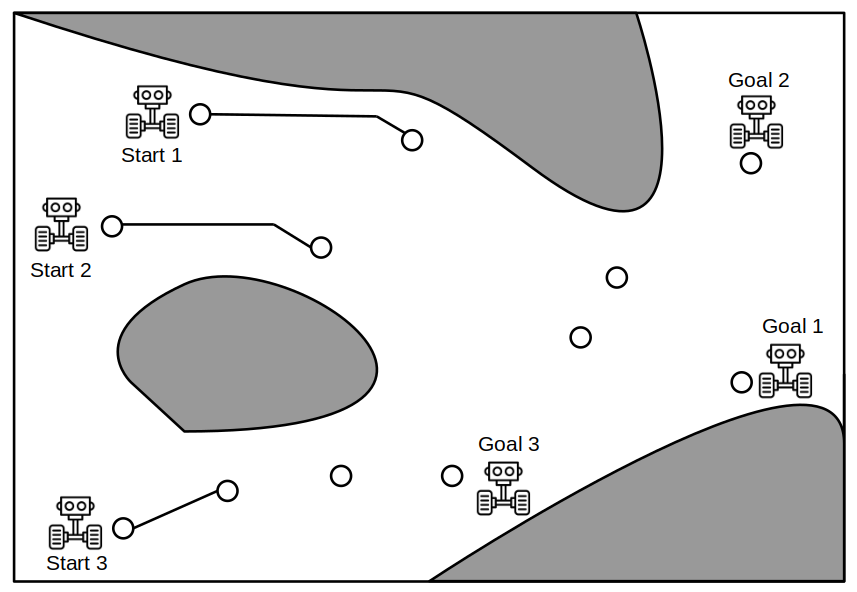}
        \caption{\footnotesize\vspace{0mm}}
    \end{subfigure}
    \begin{subfigure}[b]{0.245\linewidth}
        \includegraphics[width=1.0\linewidth]{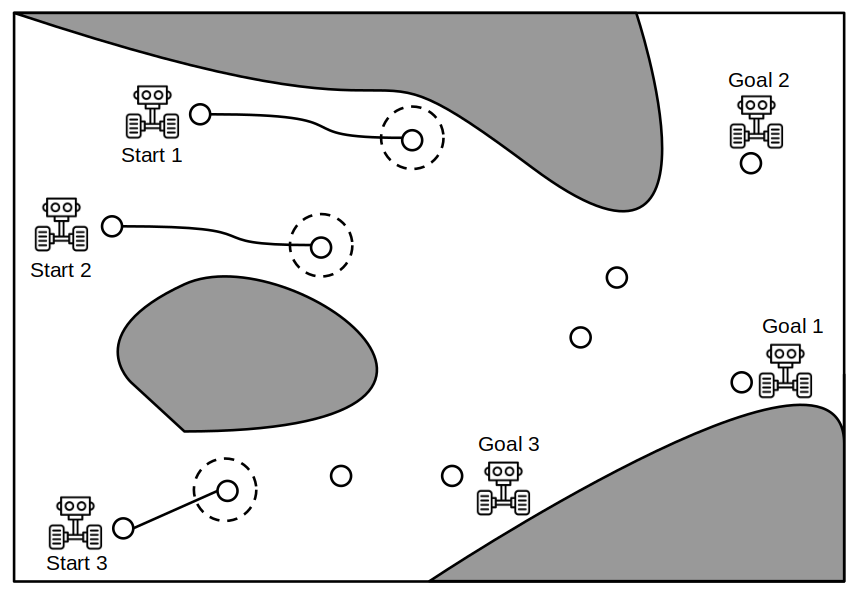}
        \caption{\footnotesize\vspace{0mm}}
    \end{subfigure}
    \begin{subfigure}[b]{0.245\linewidth}
        \includegraphics[width=1.0\linewidth]{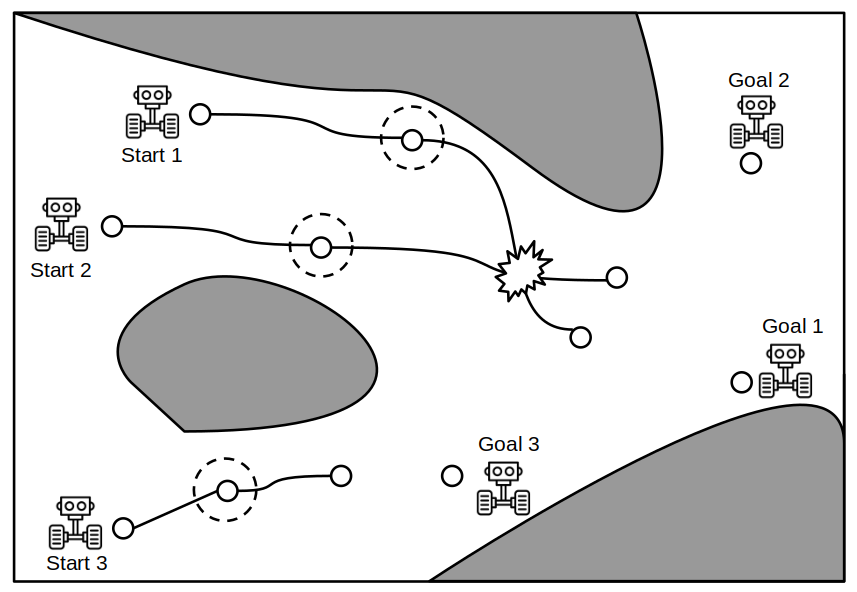}
        \caption{\footnotesize\vspace{0mm}}
    \end{subfigure}
    \caption{{\footnotesize{K-ARC's process in constructing the solution: (a) Each robot builds their individual roadmap and finds an initial kinematic path where the paths are divided into segments with equal number of timesteps. (b) We start in the first segment. (c) We convert the kinematic path into kinodynamically feasible trajectory. (d) We repeat the process for the second segment, now there is a conflict between robot 1 and robot 2's path, which we will the proceed to solve through a multi-robot planner.}      
        \vspace{0mm}}}
    \label{fig:overview}
\end{figure*}

 In this section, we first give an overview of the high-level structure of our algorithm, and then describe in detail its different planning stages. 

\subsection{Overview}

Our algorithm (Algorithm \ref{alg:K-ARC}) starts by finding the initial paths for each robot (line 3) using any kinematic sampling-based method. These paths will not used as the final solutions but rather as guidance for the kinodynamic solutions.

The individual paths are then divided into $m$ equally sized segments for each robot where we obtain a goal configuration for each segment (line 6-9), i.e., for a path consisting of a sequence of configurations for robot $r_i$, \( \{c_{i,1}, c_{i,2}, \dots, c_{i,T}\} \), the length of each path segment will be $\ceil{\frac{T}{m}}$ and the set of goals will be $G_i = \{x_{i, \ceil{\frac{T}{m}}}, x_{i, 2 \ceil{\frac{T}{m}}}, ..., x_{i, T}\}$ where each state in the set consists of its corresponding configuration. In addition, we also obtain and keep track of the initial kinematic path segment (line 10). 

For each iteration of the $m$ segments, we first define a single-robot kinodynamic planning problem (line 14). We define a query where the start will be the initial start for the first iteration and the last state from the previous segment for later iterations. The goal will then be from the set $G_i$ as previously defined, where a local tolerance will be set. We then proceed to find the kinodynamic trajectory segment for each robot through a single robot trajectory optimizer where the initial kinematic path segment will be used as the input (line 17). 

The trajectory segments are then checked for inter-robot collisions (line 20). The collisions are resolved by first defining a multi-robot kinodynamic planning problem. The query for each robot will then be the same as the single-robot query, and we are free to choose any multi-robot kinodynamic planner to solve the problem.

After the solution for a segment is found, we obtain the goal for each robot from the solution and set it as the start for the next segment (line 23). The solutions are then sequentially constructed, until the last segment where we arrive at the final goal within a tolerance. Through such sequential construction of the solutions, we make sure that in each segment, the robot would start from the last state that it arrives from the previous segment, regardless of whether this last state is equal to the goal state that we defined. Thus this is able to resolve the aforementioned issue that would arise from a direct extension from ARC. 

    

\begin{algorithm}[h!]
    \caption{Kinodynamic Adaptive Robot Coordination (K-ARC)}\label{alg:K-ARC}
    \label{alg:overview}		
    \begin{algorithmic}[1]
        \INPUT A kinodynamic planning problem with an environment $\env$, a set of robots $\robots$, a set of queries $\query$.
        \OUTPUT A set of valid, kinodynamically feasible paths $\paths_k$.
        \State $\paths\leftarrow\emptyset$ \label{alg:empty_paths_1}
        \For {each robot $r_i \in \robots$ } \label{alg:initial_paths}
        \State $p_i\leftarrow$\texttt{MotionPlanning}($\env,\{r_i\},\{q_i\}$)
        \State $\paths\leftarrow \paths\cup\{p_i\}$   
        \EndFor \label{alg:empty_paths_2}
            \State $m \leftarrow$ \texttt{number of path segments}
        \For {each robot $r_i \in \robots$} \label{alg:segment}
        \State $p_{r_i} \leftarrow \paths[i]$   \Comment{retrieve the kinematic path for robot $i$}
        \State $G_{r_i} = \{g_1, g_2, \dots, g_m\}$
        \State $\paths_{r_i} = \{p_1, p_2, \dots, p_m\} \leftarrow \texttt{SegmentPath}(p_{r_i}, m)$

        \EndFor \label{alg:segment_paths}
            \State $\paths_k \leftarrow \emptyset$
            \For{$j \in m$}
            \State $\paths_i \leftarrow \emptyset$
            \For {each robot $r_i \in \robots$}
            \State $ q_{local} \leftarrow (g_{last}, \textsf{g}_j) \text{ where } \textsf{g}_j  \text { is a region centered around $G_{r_i}$[j] }  $ 
            \State $\tau_{ij}\leftarrow$\texttt{KinodynamicPlanning}($\env,\{r_i\},\{q_{local}\}, \paths_{r_i}[j]$)
            \State $\paths_j \leftarrow \paths_j \cup \{\tau_{ij}\} $
            \EndFor 
            \textcolor{black}{
        \State {$\conflict$ = \texttt{FindConflict($\paths_j$)}} \label{alg:first_conflict}
            \State \texttt{$\env'$,$\robots'$,$\query'$ = CreateSubProblem($\conflict,\paths_j,\env$)}
            \State \texttt{$\paths'_j$ = SolveSubProblem($\env', \robots', \query'$)}
            \If {$\paths'_j != \emptyset$} \Comment{conflict resolved}
                \State \texttt{UpdateSolution($\paths_k,\paths'_j$)} 
                \State \texttt{UpdateLastGoals($\paths'_j$)}
            \Else  \Comment{if conflict not resolved,}
                \State \Return $\emptyset$ \Comment{return empty solution}
            \EndIf 
            \EndFor
            }
        \State\Return $\paths_k$\label{alg:k-arc}			
    \end{algorithmic} 
\end{algorithm}	

\subsection{Single robot trajectory Optimizer}

The initial path constructed in each segment can be used as a reference for the trajectory optimizer to find the kinodynamic path. The optimization equation for each robot $r_i$ is in Equation (\ref{eq:optimizer}). 

\begin{equation}
    \label{eq:optimizer}
    \begin{aligned}
        & \underset{X, U, \Delta t}{\text{minimize}}
        & & \sum_{k=1}^{T}J(x_{i,k}, u_{i,k}) \\
        & \text{subject to}
        & &      x_{i, 1} = x_{i, s} \quad \text{(1)}\\ 
        & & &    x_{i, T} \in X_{i, g} \quad \text{(2)}\\
        & & &    x_{i, k + 1} = x_{i, k} + f(x_{i, k}, u_{i, k})\Delta t \quad \text{(3)}\\
        & & &    u_{i, k} \in U \quad \text{(4)}\\
        & & &    c_{i, k} \in W_{free} \text{ where } c_{i, k} \subseteq x_{i, k}  \quad \text{(5)}
    \end{aligned}
\end{equation}

The goal of the optimization is to obtain a kinodynamic path from the non-kinodynamic path found by a regular motion planner, so we set the dynamic function as a hard constraint (3). We require the control inputs to be within the allowed set of controls and the configuration of each state to be within the free space of the environment(4)(5).

We set our cost function for a given timestep $ k $ to be $ J(x_{i, k}, u_{i, k}) = \beta_1 \|\mathbf{u}_{i, k}\|^2 + \Delta t$ to minimize the control input and $\Delta T$. The main purpose of the trajectory optimizer is to make sure the paths are dynamically feasible, so the first term in reality can be formulated differently, as long as (3) holds.

On the other hand, we want to minimize the total amount of time the output trajectory takes. Since the number of decision variables is fixed and we cannot directly set the number as a decision variable, we can instead minimize $\Delta T$ and set it as a decision variable. This is similarly addressed in \cite{Moldagalieva}. 

We also note that this optimizer is applied as an intermediate step during the planning process (Algorithm \ref{alg:K-ARC}, line 15) as opposed to the final step. Doing so results in a smaller optimization problem and allows for faster convergence.





\subsection{Subproblem Construction and Planning}

The kinodynamic paths obtained from the trajectory optimizer are checked for conflicts. A conflict is defined as a tuple of $(x_{i, k}, x_{j, k})$ for robot $r_i$ and $r_j$ given a timestep $k$. Given such a conflict, we define a local subproblem $(\robots',\query')$ around the conflict (Alg. \ref{alg:solve_subproblem})
where $\robots'=r_i\cup r_j$ merges the involved robots.
The local query $\query'$ defines the local start and goal state for each robot and is obtained by taking the start state and goal state for the segment where the conflict occurs. 

Again, to make sure the local subproblem is solvable, we define a tolerance for the goal state, where we then attempt to solve the local subproblem with a MRKP solver. The conflict resolution follows a similar process as in ARC \cite{solis-arc}, as shown in Alg. \ref{alg:solve_subproblem} (the differences from ARC are highlighted), although several modifications are made due to the added complexity on the robots' movement.

First, we do not limit planning to a local environment, as smaller environments overly restrict the planning space for conflict resolution, and allow the planner to operate in the entire environment. Second, if a solution is not found $(\env, \robots',\query')$, we adapt the subproblem by setting the start query to the robot's previous segment start and the goal query to its next segment goal (line 10), as opposed to in ARC where the queries are obtained by small incremental expansions. These changes address the added complexity of kinodynamic planning, where larger state spaces enable a higher chance of finding a feasible trajectory, especially for robots with highly constrained motions.

Lastly, in subproblem planning (line 4), the following hierarchy of solvers are used:
\begin{itemize}
    \item Prioritized trajectory optimization 
    \item Decoupled Kinodynamic RRT
    \item Composite Kinodynamic RRT
\end{itemize}

Initially, conflicts are resolved through prioritized trajectory optimization. For robots $\{r_1, r_2, ..., r_k\}$, we first find the kinematic paths through a group planner. The paths are then sequentially optimized, i.e., $r_1$'s path will be directly fed into Equation (\ref{eq:optimizer}), $r_2$'s path will be fed into Equation (\ref{eq:optimizer}) with the added constraint that each state in $r_2$'s path is a certain distance away from each state in $r_1$'s path, etc. 

Subsequent solvers are sampling-based to provide better exploration if the prioritized trajectory optimization fails. Here, we employ two multi-robot variants of Kinodynamic RRT \cite{lavalle-kino-rrt}.



	\begin{algorithm}[t]
		\caption{SolveSubProblem}\label{alg:solve_subproblem}
		\begin{algorithmic}[1]
			\INPUT \textcolor{blue}{A subproblem with environment $\env$}, a set of robots $\robots$, a set of queries $\query'$.
                \textcolor{blue}{A set of MRKP solvers $\mathcal{S}$}.
			\OUTPUT A set of valid local paths $\paths'$.
            \State $\paths'\leftarrow\emptyset$
            \While{$\paths'== \emptyset$}  \label{alg:while}		
                \For {$s$ \textbf{in} $\mathcal{S}$ } \label{alg:solvers_subproblem}       
        			\State \textcolor{blue}{{$\paths'\leftarrow$\texttt{SolveMRKP($s,\env',\robots',\query'$)}}} \label{alg:solve_local_subproblem}
        			\If{$\paths' \neq \emptyset$} \Comment{subproblem solved}
                        \State{}
        			    \Return {$\paths'$}
        			\EndIf
                \EndFor
                \If {$\env' \neq \env$ \textbf{or} $\query' \neq \query$}
                    \State \textcolor{blue}{{\texttt{AdaptSubProblem($\env',\robots',\query'$)}}}\label{alg:expand_subproblem}
                    
                \Else 
                    \State {}
                    \Return $\emptyset$  \label{alg:fail_subproblem} \Comment{if subproblem not solved,}
                \EndIf \Comment{return empty local solution}
            \EndWhile
			
		\end{algorithmic}
    
	\end{algorithm}

\subsection{Design choices}
We recognize that several components in our method can be approached differently, and here we discuss several components where other options are available.

1) \textit{Initial pathfinding:} In Algorithm \ref{alg:K-ARC}, line 3, we plan paths for the robots individually, without checking for conflicts. The purpose of this is to quickly find feasible paths that can be used as rough guidance. Alternatively, we can choose to run ARC instead, where the resulting kinematic paths will be conflict free. Here, the tradeoff is between whether we want the initial paths to be more refined so that there is less work to do for the conflict resolution, or the other way around. In the former, we can quickly find feasible paths but need to rely more heavily on the conflict resolution. In the latter, we will spend more time on making the initial paths better for guiding the kinodynamic planner, and while we still need to check for and resolve any conflicts, the expectation is that there will be much fewer conflicts as a result.  

2) \textit{Path segmentation:} In Algorithm \ref{alg:K-ARC}, line 9-10, we segment the paths equally sized for each robot where we used a fixed number $m$, which means that the segments can be of different timesteps between different robots. We choose to do this since it is easier to keep track of the milestones that the robots achieve as we loop through the algorithm. Alternatively, we could choose a fixed number of timesteps to segment the paths, which could result in longer paths to have more segments and shorter paths to have less. Doing the segmentation this way helps synchronize the movements the robots, since the robots need to achieve the milestones for a segment at the same time. While this can be beneficial for paths that are very different in lengths, we also need to manually add waiting states for robots that arrive first at their goals for a specific segment, creating unnecessary overheads for the solution.

\section{Experiments}

In our experiments, we compare our method against one sampling-based method, K-CBS \cite{Kottinger}, and one optimization-based method, CB-MPC \cite{10611078} to highlight the strengths of using a hybrid approach. We test the methods on three different virtual scenarios and discuss the results.  

\subsection{Experiment Setup}
Our implementation of K-CBS and CB-MPC follows the same high-level algorithm framework as in the original CBS work \cite{SHARON201540} and does not include the further improvements on the framework. In our implementation of the low-level planners, K-CBS uses kinodynamic RRT \cite{lavalle-rrt} and CB-MPC uses RRT \cite{lavalle-rrt}. The two methods also differ in how their constraints are set up for conflict resolution. The constraint is represented as a tuple of a configuration and a time interval. In K-CBS, we pass this constraint to the lower level query method, which we utilize Safe Interval Path Planning (SIPP) \cite{Likhachev}. In CB-MPC, we pass this constraint to our trajectory optimizer, and it's represented as signed distance constraint, i.e., for constraint $(c, {t, t+h})$, we require the configuration to be at least one robot apart from $c$ from time $t$ to $t + h$. Our trajectory optimizers for K-ARC and CB-MPC are implemented with Casadi \cite{casadi} and solved with IPOPT \cite{ipopt}. For simplicity, we use simple polyhedrons for both our obstacle and robot models, and we calculate the shortest distances between any two objects to be used for when we set up the constraints.

All the methods are implemented in C++ in the open-source Parasol Planning Library, and the experiments are run on a machine with 32-core Intel(R) Core(TM) i9-14900K and 64 GB of RAM. Each method is given 20 trials for each scenario, with a timeout of 600 seconds. 

\subsection{Scenarios}
\begin{figure}[t]
\footnotesize
		\centering
		\begin{subfigure}[b]{0.49\linewidth}
			\includegraphics[width=\linewidth]{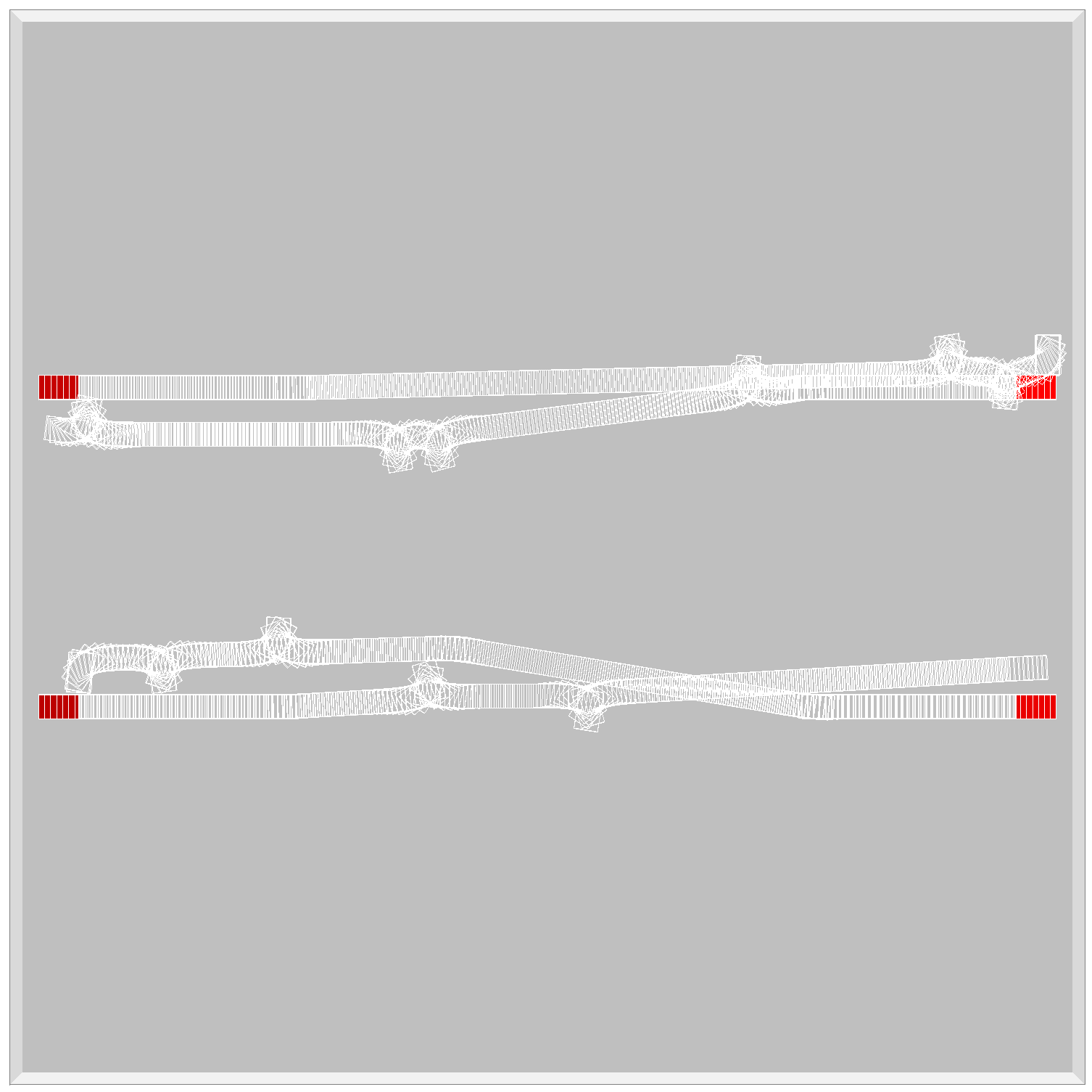}
			\caption{\footnotesize\vspace{0mm}}
		\end{subfigure}
		\begin{subfigure}[b]{0.49\linewidth}
			\includegraphics[width=\linewidth]{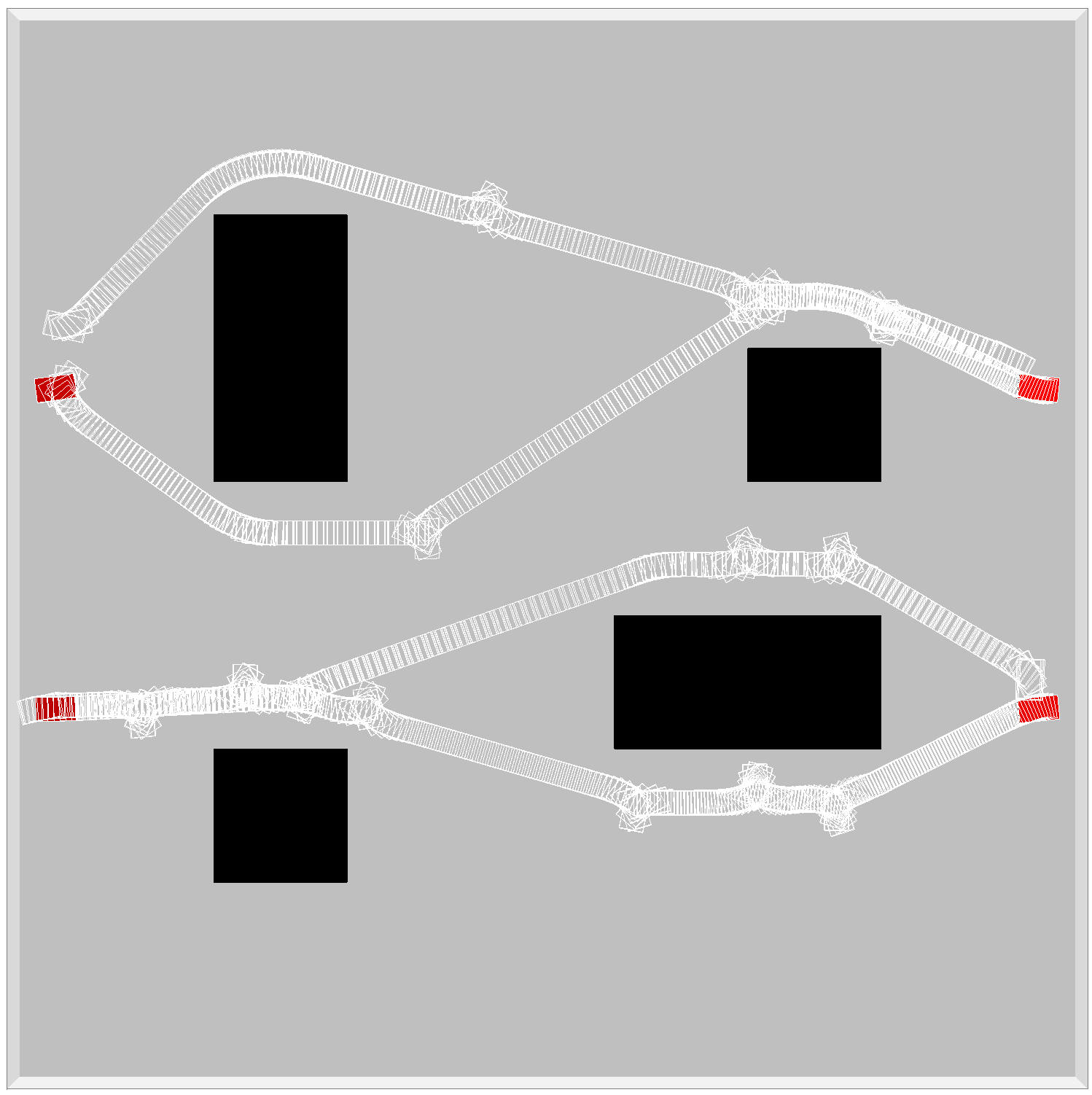}
			\caption{\footnotesize\vspace{0mm}}
		\end{subfigure}
        
        \vspace{0.1in}
		\begin{subfigure}[b]{0.48\linewidth}
			\includegraphics[width=\linewidth]{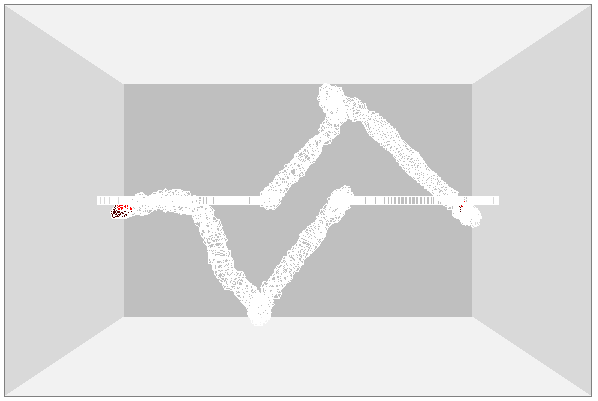}
			\caption{\footnotesize\vspace{0mm}}
		\end{subfigure}
        \vspace{0.1in}
		\begin{subfigure}[b]{0.48\linewidth}
			\includegraphics[width=\linewidth]{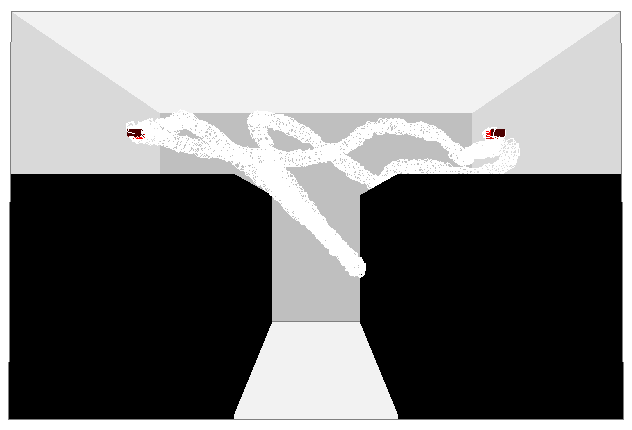}
			\caption{\footnotesize\vspace{0mm}}
		\end{subfigure}
		\caption{{\footnotesize{{The experiment environments. The paths shown in white are produced by K-ARC. (a) 4-robot open cross scenario (b) 4-robot cluttered cross scenario (c) 2-robot quadrotor scenario  (d) 2-robot quadrotor inlet scenario  }}  
        \vspace{0mm}}}
		\label{fig:scenarios}
\end{figure}

We give brief descriptions to the scenarios that we are testing the methods on.

\subsubsection{Open Cross} We examine a simple cross scenario as illustrated by Fig. \ref{fig:scenarios}(a) where robots on the same row need to swap positions in an empty environment. This is an easy scenario to test how many robots the methods are able to scale up to. In this and the following cluttered cross environment, we used both first-order and second-order unicycle robot models.

\subsubsection{Cluttered Cross} We examine a harder scenario as illustrated by Fig. \ref{fig:scenarios}(b) where the environment is filled with cluttered obstacles. The robots have the same starts and goals as in open cross. The obstacles add extra constraints to the robots movements, making this a substantially more difficult problem, and are used to test how the methods are able to handle complex environments. 

\subsubsection{Quadrotor Cross} We examine a scenario as illustrated by Fig. \ref{fig:scenarios}(c) where multiple 6-DOF quadrotor robots need to move across in an empty environment. This scenario is set up to test the methods' ability to handle robots with complex constraints and high-dimensional state space. We also relax the goal tolerance for this scenario for the methods to be able to handle larger number of robots.

\subsubsection{Quadrotor Inlet} We examine a highly-constrained scenario as illustrated by Fig. \ref{fig:scenarios}(d) where 2 quadrotor robots need to switch positions in an environment with narrow passages and an inlet to allow extra space for the robots to move. The tight obstacle constraints, combined with the complex dynamic constraints of the quadrotor robots, make this scenario particularly challenging, where the methods can only handle 2 robots.

\subsection{Robot Models}
We briefly describe the robot models used in the experiments.

\paragraph{First-Order Unicycle} The first-order unicycle model assumes the robot can directly control its linear velocity \( v \) and angular velocity \( \omega \). The robot's state consists of its position and orientation, \( [x, y, \theta] \). 

\paragraph{Second-Order Unicycle} The second-order unicycle model includes dynamics for acceleration. The state is \( [x, y, \theta, v, \omega] \), where \( v \) and \( \omega \) are the linear and angular velocities. The controls are linear acceleration \( a \) and angular acceleration \( \alpha \). 

\paragraph{Second-Order Quadrotor} We use a second-order quadrotor model as a simplified version of the actual quadrotor dynamics. The state includes position \( [x, y, z] \) and orientation \( [\phi, \theta, \psi] \) (roll, pitch, yaw), as well as the linear velocity \( [\dot{x}, \dot{y}, \dot{z}] \) and angular velocity \( [\dot{\phi}, \dot{\theta}, \dot{\psi}] \). The controls are thrust \( T \) and torque \( [\tau_\phi, \tau_\theta, \tau_\psi] \).

\subsection{Results}

\begin{figure*}[htbp!]
\vspace{-0.4in}
\footnotesize
		\centering
		\begin{subfigure}[b]{1\linewidth}
			\includegraphics[width=\linewidth]{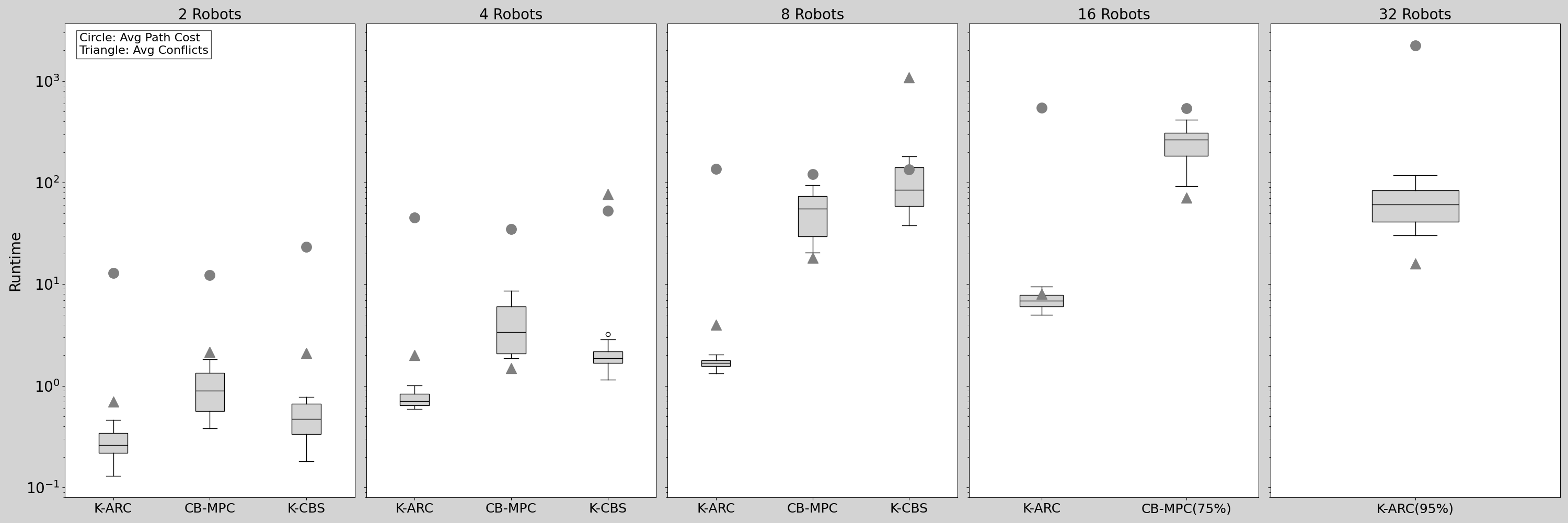}
			\caption{\footnotesize{The open cross scenario for the first order model}}
		\end{subfigure}

        \vspace{0.07in}
		\begin{subfigure}[b]{1\linewidth}
			\includegraphics[width=\linewidth]{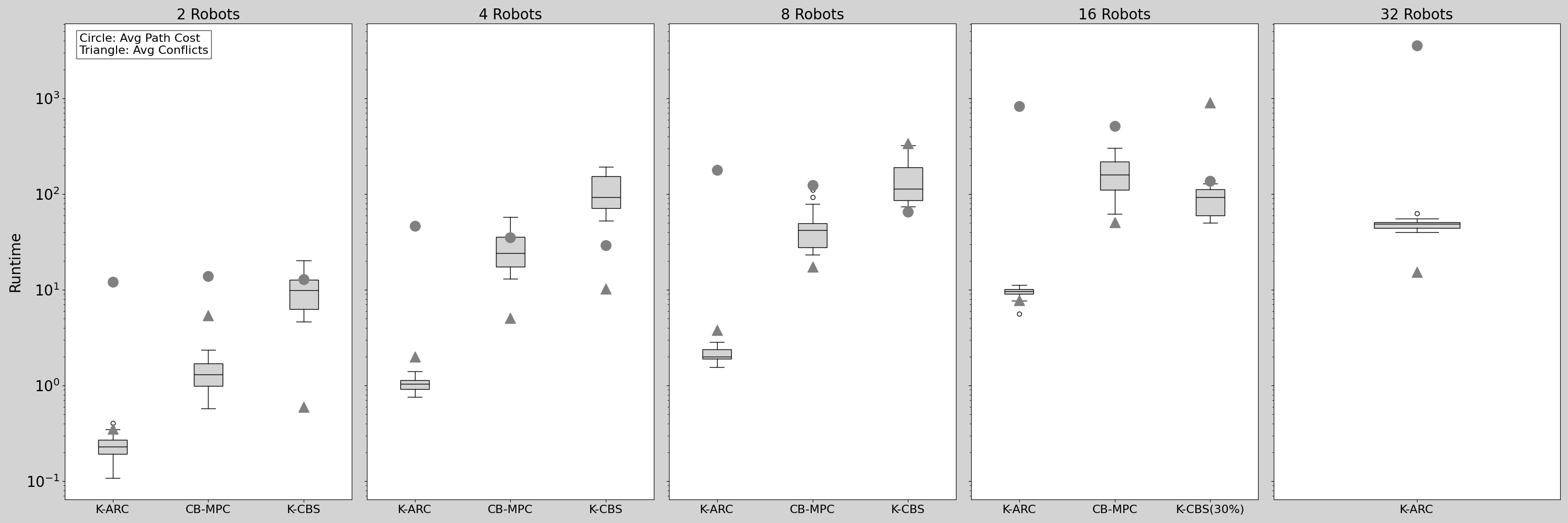}
			\caption{\footnotesize{The open cross scenario for the second order model}}
		\end{subfigure}

        \vspace{0.07in}
		\begin{subfigure}[b]{0.49\linewidth}
			\includegraphics[width=\linewidth]{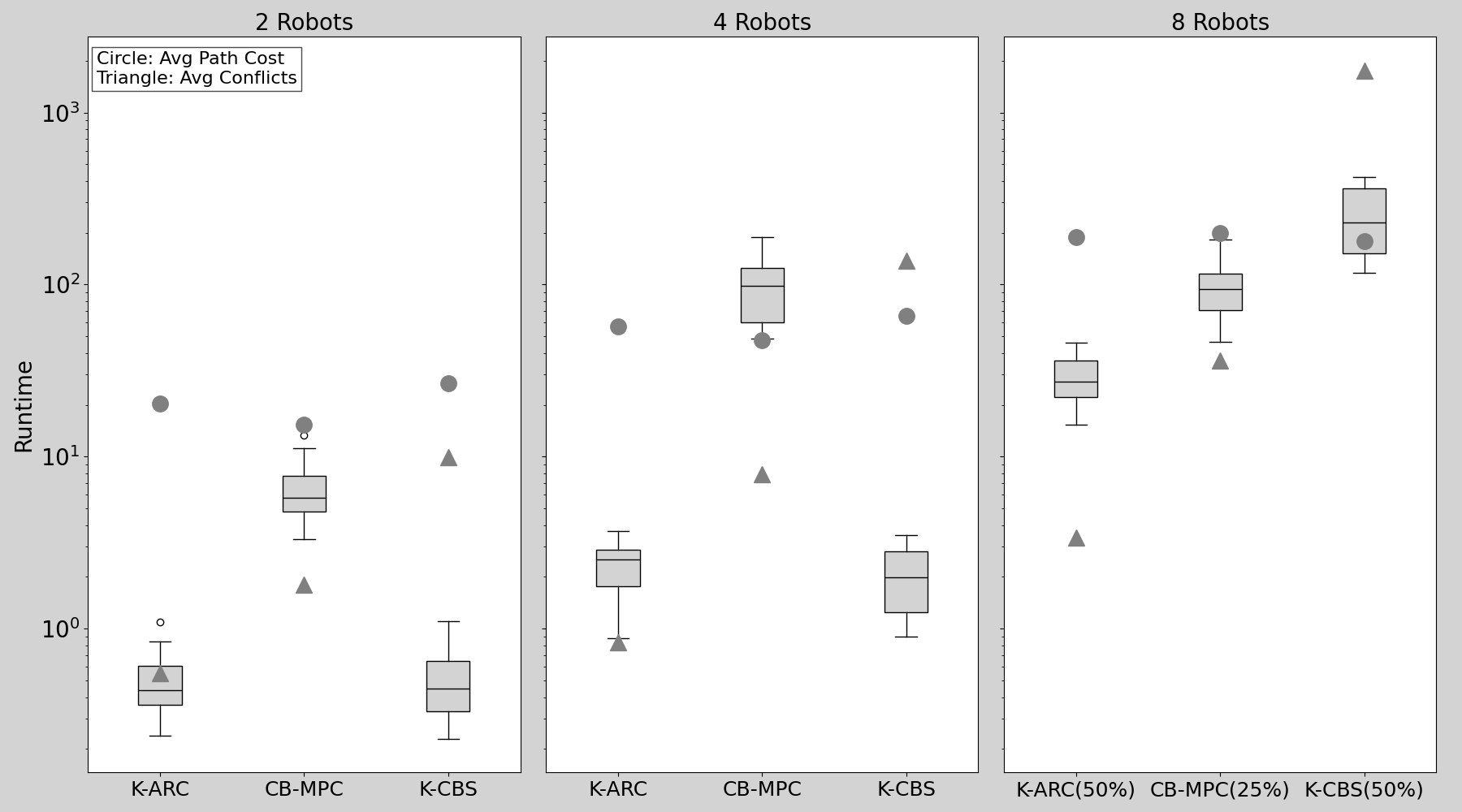}
			\caption{\footnotesize{The cluttered cross scenario for the first order model}}
		\end{subfigure}
		\begin{subfigure}[b]{0.49\linewidth}
			\includegraphics[width=\linewidth]{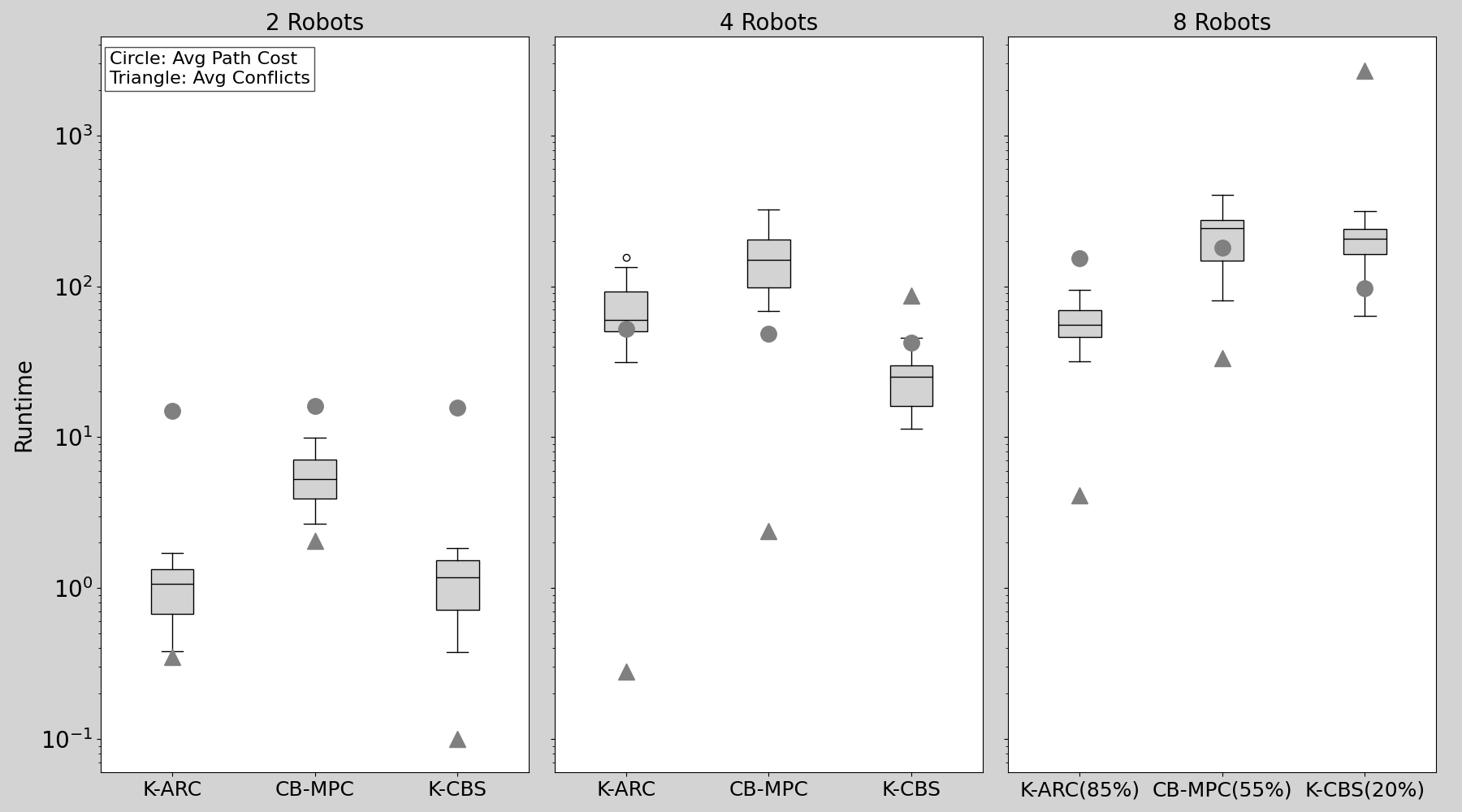}
			\caption{\footnotesize{The cluttered cross scenario for the second order model}}
		\end{subfigure}

        \vspace{0.07in}
		\begin{subfigure}[b]{0.68\linewidth}
			\includegraphics[width=\linewidth]{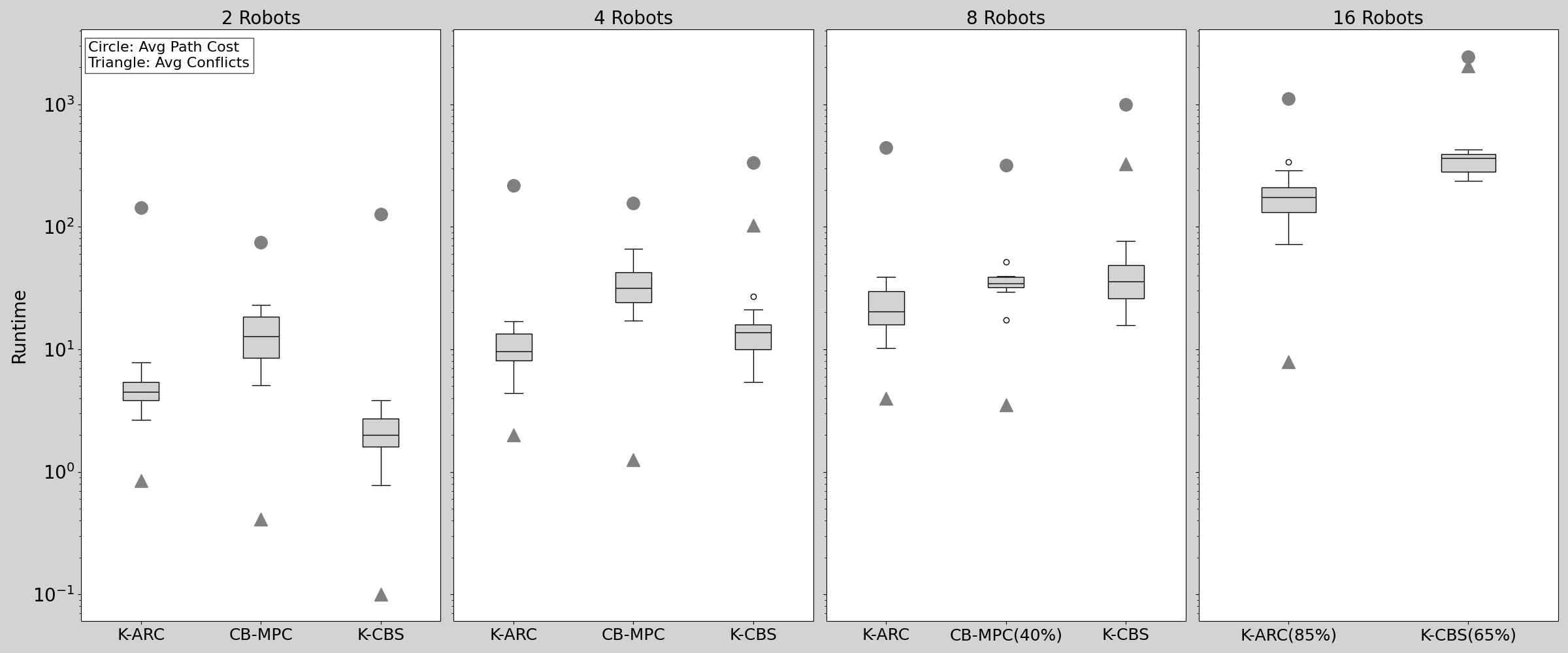}
			\caption{\footnotesize{The quadrotor cross scenario}}
		\end{subfigure}

        \caption{Results for the experiments. The percentage indicates the success rate of the method and is otherwise 100\%. }
		\label{fig:plots}
\end{figure*}

We evaluate the methods based mainly on the runtime. The runtime as shown in the tables is the amount of time in seconds a method spends in finding a solution. We also evaluate the methods on path cost as a secondary metric. We calculate the path cost as the sum of timesteps of all trajectories times the time resolution. Additionally, we show the number of conflicts each method has to resolve in each of the scenarios. Methods with the fastest runtime are bolded. 

\subsubsection{Open Cross} In this scenario to test the scalability of the methods, we find that K-ARC is able to achieve much better runtime for all numbers of robots and is the only method able to scale up to 32 robots, as shown in Figs. \ref{fig:plots}(a) and \ref{fig:plots}(b). We deem that this is mainly due to K-ARC's property to handle the conflicts as local subproblems and its approach to partially optimize trajectories. Both help reduce the complexity of the problem significantly. 

Comparing K-ARC with CB-MPC, which both use optimizations, K-ARC has consistently better runtime. Despite CB-MPC taking an MPC-based approach that uses a sliding window, the optimizer still needs to cover the entire trajectory, making it generally slower than our approach, which only optimizes a comparably shorter trajectory for every conflict resolution. 

For methods that use CBS, CB-MPC has slightly better performance than K-CBS for the first-order robot, in both runtime and path cost. While K-CBS can achieve better path cost for the second-order robot, it is unable to keep up after 16 robots in both cases. This is mainly due to the large number of constraints imposed on SIPP. This is less of an issue for CB-MPC, mainly because the constraints imposed on the optimizer are generally not too tight with the large open space. 

An interesting observation from the results is that K-CBS resolves significantly more conflicts than K-ARC and CB-MPC but still achieves comparable runtime. This difference stems from how the methods handle conflicts. K-ARC and CB-MPC use trajectory optimizers to search the continuous state space, offering more solution options but at a higher computational cost. In contrast, K-CBS relies on SIPP, which operates on a discrete graph. While this restricts solution options, it simplifies the problem and allows for faster conflict resolution, as reflected in the shorter average query times for SIPP in K-CBS.

\subsubsection{Cluttered Cross} In this scenario to test how the methods handle complex environments, we find that no methods can scale over 8 robots and have overall poorer performance compared to the open cross scenario, as shown in  Figs. \ref{fig:plots}(c) and \ref{fig:plots}(d). This is expected as this is a highly constrained scenario. However, K-ARC is still able to outperform both baseline methods in the 8-robot scenarios, whereas K-CBS has better performance in the 2-robot and 4-robot scenario. 

On the other hand, CB-MPC has generally poorer performance than both K-CBS and K-ARC. K-ARC also does not have as good performance as in the open-cross scenario. This is a result of the added complexity of obstacle constraints that affects optimization-based and sampling-based approaches differently. In trajectory optimization, obstacle constraints are applied to every state, which can create extremely tight restrictions. In contrast, sampling-based methods inherently handle obstacles, as sampled states are known to be collision free, eliminating the needs for collision checking during graph queries.   




\subsubsection{Quadrotor Cross} In this scenario to test how the methods handle complex robot dynamics, we find that only K-ARC and K-CBS are capable of planning up to 16 robots, as shown in Fig. \ref{fig:plots}(e), while CB-MPC has trouble planning for 16 robots. In the 16-robot scenario, K-ARC not only has less than half of the runtime compared to K-CBS, it also has less than half of the path cost. We recognize that this discrepancy is due to the randomness in K-CBS's purely sampling-based approach. Given the large search space of this scenario, K-CBS tends to produce poor-quality tree in the initial pathfinding stage, which impacts both its runtime and path cost. CB-MPC, on the other hand, consistently achieves better path cost than the other two methods. However, its optimization-based process soon becomes computationally expensive given the large state space of the quadrotor robot model and as the number of robots scale up.

{\subsubsection{Quadrotor Inlet} In this highly constrained scenario, we find that only K-ARC and K-CBS are able to successfully solve all the trials, as shown in Table \ref{tab:quadrotor-inlet}. In our experiments, we recognize one drawback of K-ARC is its overhead of preprocessing the conflicts, resulting in a slower runtime than K-CBS for smaller number of robots, as evident in Figs. \ref{fig:plots}(c), \ref{fig:plots}(d), and \ref{fig:plots}(e). A similar result is also seen in our previous work \cite{solis-arc}. Here, we show that K-ARC is better for a highly constrained situation like this, where K-ARC is able to quickly identify areas in the environment needed to resolve the conflicts.



\begin{table}[h!]
    \centering
    \Large
    \scalebox{.44}{
        \def\arraystretch{1.35}
        \begin{tabular}{ | c | c | r | r | r | r | r | r | r | r |}
            \hline
            \multicolumn{1}{|c|}{\multirow{2}{*}{\textbf{Robots}}} & 
            \multicolumn{1}{c|}{\multirow{2}{*}{\textbf{Model}}} & 
            \multicolumn{1}{c|}{\multirow{2}{*}{\textbf{Method}}} & 
            \multicolumn{2}{c|}{\textbf{Runtime}} &
            \multicolumn{2}{c|}{\textbf{Path Cost}} & 
            \multicolumn{2}{c|}{\textbf{Conflicts}} & 
            \multicolumn{1}{c|}{\multirow{2}{*}{\textbf{Success}}} \\ \cline{4-9}
            \multicolumn{1}{|c|}{} & 
            \multicolumn{1}{c|}{} & 
            \multicolumn{1}{c|}{} & 
            \multicolumn{1}{c|}{\textbf{Avg}} &
            \multicolumn{1}{c|}{\textbf{Std Dev}} &
            \multicolumn{1}{c|}{\textbf{Avg}} & 
            \multicolumn{1}{c|}{\textbf{Std Dev}} & 
            \multicolumn{1}{c|}{\textbf{Avg}} & 
            \multicolumn{1}{c|}{\textbf{Std Dev}} & 
            \multicolumn{1}{c|}{} \\ \hline
            
            \multirow{3}{*}{2} & \multirow{3}{*}{2nd order} & \textbf{K-ARC}  & \textbf{69.34}   & 81.98    & 168.5   & 40.37   & 0.56  &  0.50 &  100\%     \\ \cline{3-10}            
            &  & CB-MPC  & -  & -   & -   & -  &  -  &	-  & 0\%    \\ \cline{3-10}  
            &  & K-CBS   & 144.71  & 116.66   & 221.22   & 84.07  &  2.28  &  1.34  & 90\%     \\ \Xhline{2.5\arrayrulewidth}
            \hline
        \end{tabular}
    }
    \caption{\small{Results for he quadrotor inlet scenario}}
    \label{tab:quadrotor-inlet}
\end{table}
} 

\section{Conclusions}
In this paper, we presented Kinodynamic Adaptive Robot Coordination (K-ARC), a multi-robot kinodynamic planning algorithm, as an extension to a previously proposed multi-robot motion planning algorithm. K-ARC benefits from utilizing both sampling-based and optimization-based technique in different stages of the planning process, and is shown to achieve better performance compared to baseline methods found in current literature. Through this work, we recognize that a combination of different classes of approaches leads to an overall better algorithm. In the future, we plan to integrate learning-based approach into our work, specifically integrating the idea of ARC into the work of multi-agent reinforcement learning (MARL).

 

	
	\bibliographystyle{ieeetr}
	\bibliography{main}

\begin{thebibliography}{10}

\bibitem{lavalle-kino-rrt}
S.~LaValle and J.~Kuffner, ``Randomized kinodynamic planning,'' in {\em Proceedings 1999 IEEE International Conference on Robotics and Automation (Cat. No.99CH36288C)}, vol.~1, pp.~473--479 vol.1, 1999.

\bibitem{SHARON201540}
G.~Sharon, R.~Stern, A.~Felner, and N.~R. Sturtevant, ``Conflict-based search for optimal multi-agent pathfinding,'' {\em Artificial Intelligence}, vol.~219, pp.~40--66, 2015.

\bibitem{Moldagalieva}
A.~Moldagalieva, J.~Ortiz-Haro, M.~Toussaint, and W.~Hönig, ``db-cbs: Discontinuity-bounded conflict-based search for multi-robot kinodynamic motion planning,'' in {\em 2024 IEEE International Conference on Robotics and Automation (ICRA)}, pp.~14569--14575, 2024.

\bibitem{Juncheng}
J.~Li, M.~Ran, and L.~Xie, ``Efficient trajectory planning for multiple non-holonomic mobile robots via prioritized trajectory optimization,'' {\em IEEE Robotics and Automation Letters}, vol.~6, no.~2, pp.~405--412, 2021.

\bibitem{10611078}
A.~Tajbakhsh, L.~T. Biegler, and A.~M. Johnson, ``Conflict-based model predictive control for scalable multi-robot motion planning,'' in {\em 2024 IEEE International Conference on Robotics and Automation (ICRA)}, pp.~14562--14568, 2024.

\bibitem{Kottinger}
J.~Kottinger, S.~Almagor, and M.~Lahijanian, ``Conflict-based search for multi-robot motion planning with kinodynamic constraints,'' in {\em 2022 IEEE/RSJ International Conference on Intelligent Robots and Systems (IROS)}, pp.~13494--13499, 2022.

\bibitem{solis-arc}
I.~Solis, J.~Motes, M.~Qin, M.~Morales, and N.~M. Amato, ``Adaptive robot coordination: A subproblem-based approach for hybrid multi-robot motion planning,'' {\em IEEE Robotics and Automation Letters}, vol.~9, no.~8, pp.~7238--7245, 2024.

\bibitem{lozano-plan}
T.~{Lozano-P\'{e}rez} and M.~A. Wesley, ``An algorithm for planning collision-free paths among polyhedral obstacles,'' {\em Communications ACM}, vol.~22, pp.~560--570, October 1979.

\bibitem{canny-plan}
J.~F. Canny, {\em The Complexity of Robot Motion Planning}.
\newblock Cambridge, MA: MIT Press, 1988.

\bibitem{kavraki-prm}
L.~Kavraki, P.~Svestka, J.-C. Latombe, and M.~Overmars, ``Probabilistic roadmaps for path planning in high-dimensional configuration spaces,'' {\em IEEE Transactions on Robotics and Automation}, vol.~12, no.~4, pp.~566--580, 1996.

\bibitem{lavalle-rrt}
S.~LAVALLE, ``Rapidly-exploring random trees : a new tool for path planning,'' {\em Research Report 9811}, 1998.

\bibitem{karaman}
S.~Karaman and E.~Frazzoli, ``Optimal kinodynamic motion planning using incremental sampling-based methods,'' in {\em 49th IEEE Conference on Decision and Control (CDC)}, pp.~7681--7687, 2010.

\bibitem{webb-kino-rrt*}
D.~J. Webb and J.~van~den Berg, ``Kinodynamic rrt*: Optimal motion planning for systems with linear differential constraints,'' 2012.

\bibitem{hauser-kino-rrt}
K.~Hauser and Y.~Zhou, ``Asymptotically optimal planning by feasible kinodynamic planning in a state–cost space,'' {\em IEEE Transactions on Robotics}, vol.~32, no.~6, pp.~1431--1443, 2016.

\bibitem{chomp}
M.~Kalakrishnan, S.~Chitta, E.~Theodorou, P.~Pastor, and S.~Schaal, ``Stomp: Stochastic trajectory optimization for motion planning,'' in {\em 2011 IEEE International Conference on Robotics and Automation}, pp.~4569--4574, 2011.

\bibitem{stomp}
N.~Ratliff, M.~Zucker, J.~A. Bagnell, and S.~Srinivasa, ``Chomp: Gradient optimization techniques for efficient motion planning,'' in {\em 2009 IEEE International Conference on Robotics and Automation}, pp.~489--494, 2009.

\bibitem{schulman}
J.~Schulman, Y.~Duan, J.~Ho, A.~X. Lee, I.~Awwal, H.~Bradlow, J.~Pan, S.~Patil, K.~Goldberg, and P.~Abbeel, ``Motion planning with sequential convex optimization and convex collision checking,'' {\em The International Journal of Robotics Research}, vol.~33, pp.~1251 -- 1270, 2014.

\bibitem{Cain}
B.~Cain, M.~Kalaitzakis, and N.~Vitzilaios, ``Mk-rrt*: Multi-robot kinodynamic rrt trajectory planning,'' in {\em 2021 International Conference on Unmanned Aircraft Systems (ICUAS)}, pp.~868--876, 2021.

\bibitem{ChenAAAI21s2m2}
J.~Chen, J.~Li, C.~Fan, and B.~Williams, ``Scalable and safe multi-agent motion planning with nonlinear dynamics and bounded disturbances,'' in {\em Proceedings of the 35th AAAI Conference on Artificial Intelligence (AAAI)}, pp.~11237--11245, 2021.

\bibitem{honig}
W.~Hönig, J.~A. Preiss, T.~K.~S. Kumar, G.~S. Sukhatme, and N.~Ayanian, ``Trajectory planning for quadrotor swarms,'' {\em IEEE Transactions on Robotics}, vol.~34, no.~4, pp.~856--869, 2018.

\bibitem{Likhachev}
M.~Phillips and M.~Likhachev, ``Sipp: Safe interval path planning for dynamic environments,'' in {\em 2011 IEEE International Conference on Robotics and Automation}, pp.~5628--5635, 2011.

\bibitem{casadi}
J.~Andersson, J.~Gillis, G.~Horn, J.~Rawlings, and M.~Diehl, ``Casadi: a software framework for nonlinear optimization and optimal control,'' {\em Mathematical Programming Computation}, vol.~11, 07 2018.

\bibitem{ipopt}
L.~T. Biegler and V.~M. Zavala, ``Large-scale nonlinear programming using ipopt: An integrating framework for enterprise-wide dynamic optimization,'' {\em Comput. Chem. Eng.}, vol.~33, pp.~575--582, 2009.

\end{thebibliography}

\end{document}